\documentclass{article}

\usepackage[dvipsnames]{xcolor} 
\usepackage[pagebackref=true,breaklinks=true,colorlinks,bookmarks=false]{hyperref}

\usepackage[preprint]{neurips_2025}
\usepackage[misc]{ifsym}

\usepackage[utf8]{inputenc} 
\usepackage[T1]{fontenc}    
\usepackage{hyperref}       
\usepackage{url}            
\usepackage{booktabs}       
\usepackage{amsfonts}       
\usepackage{nicefrac}       
\usepackage{microtype}      
\usepackage{xcolor}         

\usepackage{amsmath}
\usepackage{amssymb}
\usepackage{mathtools}
\usepackage{amsthm}

\usepackage{pifont} 
\usepackage{multirow, overpic, textpos, array}
\usepackage{makecell}

\usepackage{color, colortbl}
\definecolor{Gray}{gray}{0.95}
\definecolor{lightgray}{gray}{0.5}
\definecolor{light}{gray}{0.96}
\usepackage{nicematrix}
\usepackage{wrapfig}
\usepackage{caption}

\newlength\savewidth\newcommand\shline{\noalign{\global\savewidth\arrayrulewidth
  \global\arrayrulewidth 1pt}\hline\noalign{\global\arrayrulewidth\savewidth}}

\newcommand\blfootnote[1]{\begingroup\renewcommand\thefootnote{}\footnote{#1}\addtocounter{footnote}{-1}\endgroup}

\newcommand{\TableSystemComparison}{
\begin{table}[b!]
    \caption{System-level comparison on class-conditional generation over 50K samples on $256\times 256$ ImageNet benchmark. Note that \#params in the table only counts in AR model parameters and our tokenizer is with 300M parameters. MAR is difficult to categorize into mask-based or tailored AR methods.}\label{tab:system_comparison}

    \renewcommand{\arraystretch}{1.05}
    \setlength{\tabcolsep}{7.4pt}
    \centering
    \small
    
    \begin{tabular}{c|l|c|cc|cc}
        type & method & \#params & FID$\downarrow$ & IS$\uparrow$ & Prec$\uparrow$ & Rec$\uparrow$ \\\shline
        \hline
        diffusion
         & DiT-XL~\cite{dit} & {675M} & {2.27} & {278.2} & {0.83} & {0.57} \\
         & SiT-XL~\cite{sit} & {675M} & {2.06} & {270.3} & {0.82} & {0.59} \\
        \hline
        mask-based & MaskGIT~\cite{maskgit} & {227M} & {6.18} & {182.1} & {0.80} & {0.51} \\
         & TiTok-S-128~\cite{titok} & {287M} & {1.97} & {281.8} & {-} & {-} \\
         & MAR-L~\cite{mar} & {479M} & {1.78} & {296.0} & {0.81} & {0.60} \\
         & MAR-H~\cite{mar} & {943M} & {1.55} & {303.7} & {0.81} & {0.62} \\
        \hline
        tailored AR
         & VAR-d20~\cite{var} & {600M} & {2.57} & {302.6} & {0.83} & {0.56} \\
         & VAR-d24~\cite{var} & {1.0B} & {2.09} & {312.9} & {0.82} & {0.59} \\
         & VAR-d30~\cite{var} & {2.0B} & {1.92} & {323.1} & {0.82} & {0.59} \\
         & RAR-L~\cite{rar} & {461M} & {1.70} & {299.5} & {0.81} & {0.60} \\
         & RAR-XL~\cite{rar} & {955M} & {1.50} & {306.9} & {0.80} & {0.62} \\
         & RandAR-L~\cite{randar} & {343M} & {2.55} & {288.82} & {0.81} & {0.58} \\
         & RandAR-XL~\cite{randar} & {775M} & {2.22} & {314.21} & {0.80} & {0.60} \\
         & RandAR-XXL~\cite{randar} & {1.4B} & {2.15} & {321.97} & {0.79} & {0.62} \\
         & DART-FM~\cite{dart} & {820M} & {3.82} & {263.8} & {-} & {-} \\
         & CausalFusion-XL~\cite{causalfusion} & {676M} & 1.77 & 282.3 & {0.82} & {0.61} \\
        \hline
        vanilla AR
         & LlamaGen-L~\cite{llamagen} & {343M} & {3.07} & {256.06} & {0.83} & {0.52} \\
         & LlamaGen-XL~\cite{llamagen} & {775M} & {2.62} & {244.08} & {0.80} & {0.57} \\
         & LlamaGen-XXL~\cite{llamagen} & {1.4B} & {2.34} & {253.90} & {0.80} & {0.59} \\
         & IBQ-XL~\cite{ibq} & {1.1B} & {2.14} & {278.99} & {0.83} & {0.56} \\
         & IBQ-XXL~\cite{ibq} & {2.1B} & {2.05} & {286.73} & {0.83} & {0.57} \\
         & stronger LlamaGen-L~\cite{randar} & {343M} & {2.20} & {274.26} & {0.80} & {0.59} \\
         & stronger LlamaGen-XL~\cite{randar} & {775M} & {2.16} & {282.71} & {0.80} & {0.61} \\
         \rowcolor{Gray} & \textbf{D-AR-L (ours)} & {343M} & {2.44} & {262.97} & {0.78} & {0.61} \\
         \rowcolor{Gray} & \textbf{D-AR-XL (ours)} & {775M} & {2.09} & {298.42} & {0.79} & {0.62} \\

    \end{tabular}
\end{table}
}

\newcommand{\TableTokenizerComparison}{
\begin{figure}[b]
\centering
\begin{minipage}[t]{0.5\textwidth}
    \centering
    \small
    \setlength{\tabcolsep}{4.4pt}
    \renewcommand{\arraystretch}{1.05}
    \begin{tabular}{c|ccc}
        tokenizer & \#tokens & codebook size & rFID$\downarrow$ \\
        \shline
        RQ-VAE~\cite{rqtransformer} & 256 & 16384 & 3.20 \\
        Titok-S~\cite{titok} & 128 & 4096 & 1.71 \\
        \hline
        LlamaGen~\cite{llamagen} & 256 & 4096 & 3.02 \\
        LlamaGen~\cite{llamagen} & 256 & 16384 & 2.19 \\
        \hline
        ours & 256 & 4096 & 1.84 \\
        \rowcolor{Gray} ours & 256 & 16384 & 1.58
    \end{tabular}
\end{minipage}
\hfill
\begin{minipage}[t]{0.45\textwidth}
  \vspace{-3.0em}
  \centering
\captionsetup{type=table}
\caption{
\textbf{Reconstruction results} on ImageNet validation 50K samples with 256 discrete tokens.
We also finetune our sequential diffusion tokenizer with smaller codebook size, 4096, and compare with LlamaGen tokenizer counterpart.
}
\label{tab:tokenizer_comparison}
\end{minipage}
\end{figure}
}

\newcommand{\TableTokenizerSampling}{
\begin{figure}[t]
\centering
\small
\newcolumntype{a}{>{\columncolor{Gray}}c}
\begin{minipage}[t]{0.5\textwidth}
    \centering
    \setlength{\tabcolsep}{4.4pt}
    \renewcommand{\arraystretch}{1.05}
    \begin{tabular}{c|c|c|a|c|c}
        steps & 4 & 8 & 8, Adams 2nd & 12 & 16 \\
        \shline
        rFID$\downarrow$ & 2.35 & 1.58 & 1.52 & 1.73 & 1.93
    \end{tabular}
\end{minipage}
\hfill
\begin{minipage}[t]{0.45\textwidth}
  \vspace{-2.5em}
  \centering
\captionsetup{type=table}
\caption{
\textbf{Different sampling configurations} on our sequential diffusion tokenizer. Adams 2nd refers to the two step Adams–Bashforth solver~\cite{adams2nd}, while others use Euler.
}
\label{tab:tokenizer_sampling}
\end{minipage}
\vspace{-2em}
\end{figure}
}

\title{D-AR: Diffusion via Autoregressive Models}

\author{
  Ziteng Gao \hspace{2em} Mike Zheng Shou\textsuperscript{\textbf{$\dag$}} \\
  Show Lab, National University of Singapore \\
  \texttt{gzt@outlook.com, mike.zheng.shou@gmail.com}
}

\begin{document}

\maketitle

\vspace{-1em}
\begin{center}
    \centering
    \captionsetup{type=figure}
    \includegraphics[width=0.95\linewidth]{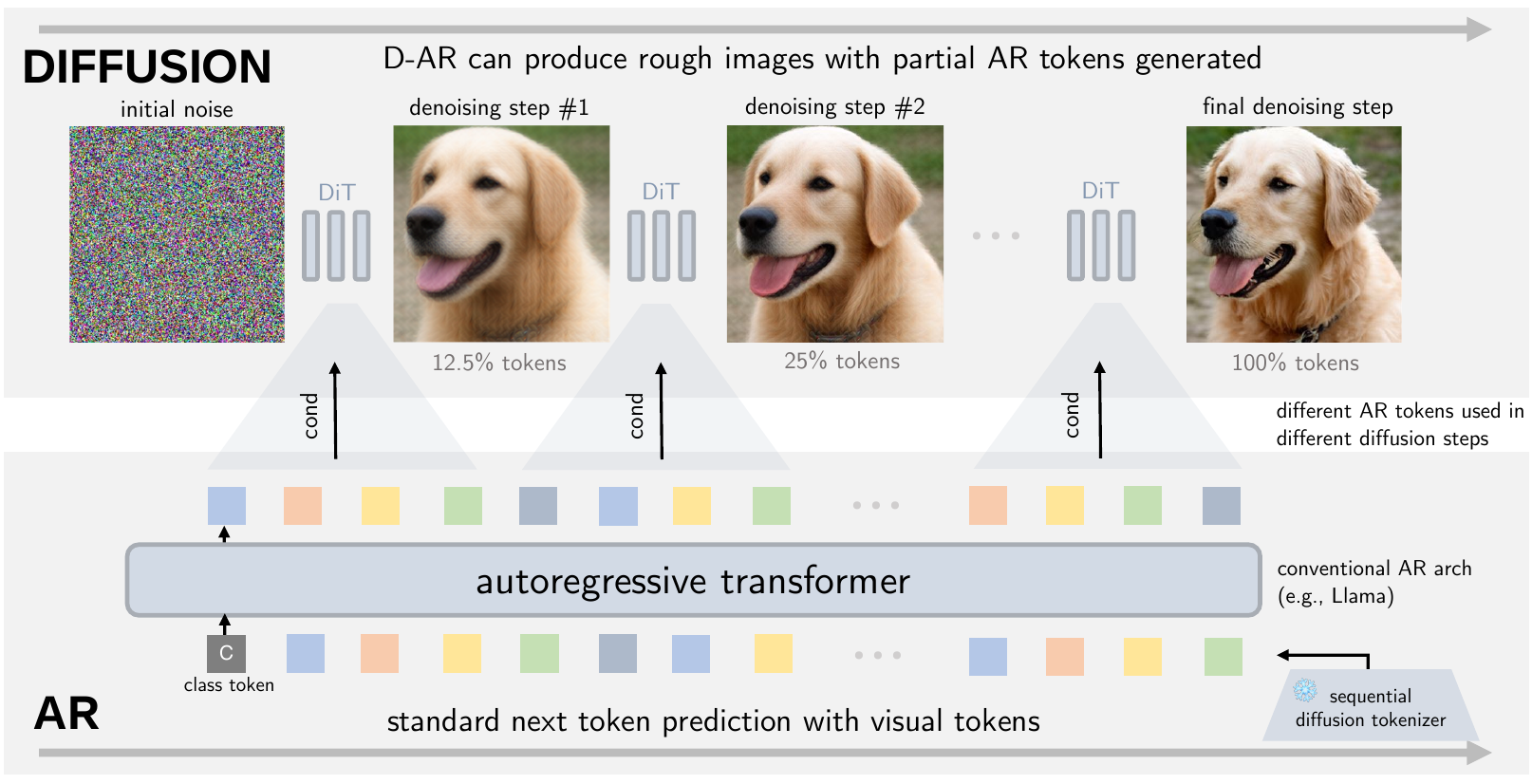}
    \caption{\textbf{Diffusion via autoregressive modeling (D-AR) framework} for visual generation. As the autoregressive transformer generates tokens, D-AR can simultaneously perform corresponding diffusion steps via token conditioning and jump-estimate target samples as rough previews effortlessly. }
    \label{fig:main_arch}
    \vspace{-0.5em}
\end{center}
\blfootnote{ \textsuperscript{\textbf{$\dag$}} Corresponding author.}

\begin{abstract}
This paper presents Diffusion via Autoregressive models (D-AR), a new paradigm recasting the image diffusion process as a vanilla autoregressive procedure in the standard next-token-prediction fashion.
We start by designing the tokenizer that converts images into sequences of discrete tokens, where tokens in different positions 
can be decoded into different diffusion denoising steps in the pixel space.
Thanks to the diffusion properties, these tokens naturally follow a coarse-to-fine order, which directly lends itself to autoregressive modeling.
Therefore, we apply standard next-token prediction on these tokens, without modifying any underlying designs (either causal masks or training/inference strategies), and such sequential autoregressive token generation directly mirrors the diffusion procedure in image space.
That is, once the autoregressive model generates an increment of tokens, we can directly decode these tokens into the corresponding diffusion denoising step in the streaming manner.
Our pipeline naturally reveals several intriguing properties, for example, it supports consistent previews when generating only a subset of tokens and enables zero-shot layout-controlled synthesis.
On the standard ImageNet benchmark, our method achieves 2.09 FID using a 775M Llama backbone with 256 discrete tokens.
We hope our work can inspire future research on unified autoregressive architectures of visual synthesis, especially with large language models.
Code and models will be available at \href{https://github.com/showlab/D-AR}{https://github.com/showlab/D-AR}.
\end{abstract}

\begin{figure}[t]
  \centering
  \includegraphics[width=0.8\textwidth]{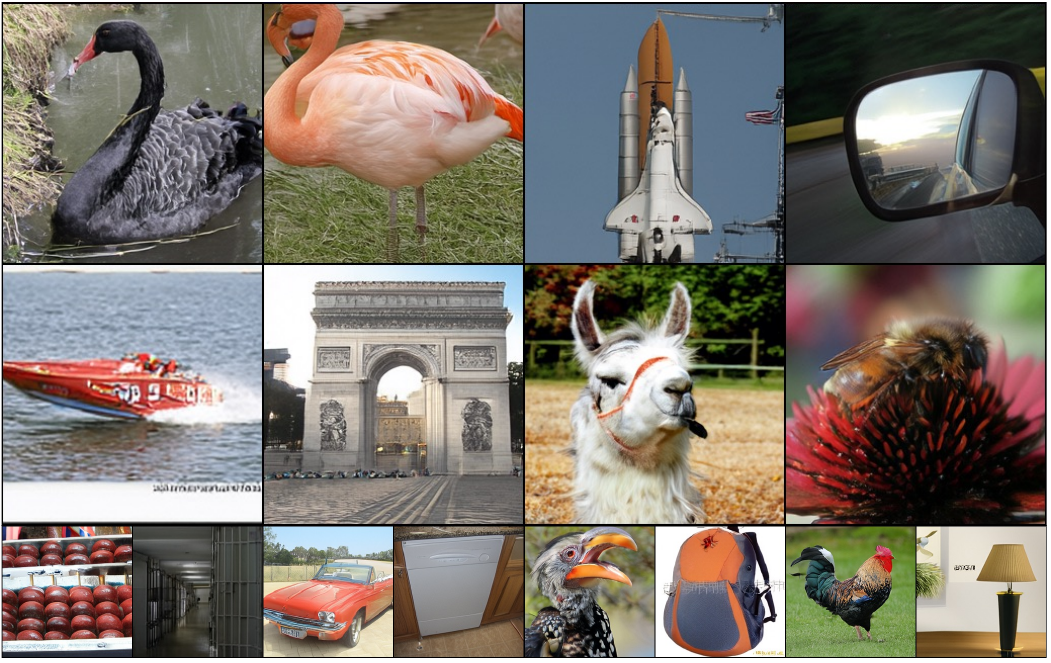}
  \caption{\textbf{Uncurated generated samples} from D-AR-XL with $256\times 256$ resolutions (CFG=4.0).}
\end{figure}

\section{Introduction}
\label{introduction}
Autoregressive models, exemplified by large language models (LLMs)~\cite{openai2024gpt4technicalreport,touvron2023llamaopenefficientfoundation,grattafiori2024llama}, have emerged as the foundation of modern NLP, achieving state-of-the-art performance with the simple next token prediction paradigm.
With their widespread adoption, the autoregressive next token prediction paradigm has established itself as the de facto standard in modern AI and this has fostered software ecosystem for optimizing such training and inference pipelines~\cite{megatron,vllm,sglang}.
The remarkable success of autoregressive models in language has also inspired exploration into visual generation tasks~\cite{vqgan,llamagen,rqtransformer}, with the broader goal of building unified frameworks of both vision and language~\cite{chameleon,transfusion,showo,janusflow}. However, unlike text, where sequential structure is naturally defined, images lack an inherently linear ordering, posing challenges for adapting such paradigm to vision modeling.
Recent studies explore different visual orderings in autoregressive modeling~\cite{var,randar,rar,xar,imagefolder}. However, these approaches typically require significant modifications to the core mechanisms, often deviating from the standard next token prediction objective.

In parallel, current vision generation pipelines are most dominated by diffusion models~\cite{ddim,ddpm,flowmatching}, which underpin several commercial systems~\cite{dalle,flux,sdxl}. Diffusion models are particularly excelling at modeling continuous signals: starting from random noise, they iteratively refine input through denoising to produce high-quality images.
However, despite impressive visual fidelity, the diffusion sampling process, operating in a dense manner, is inherently slow, especially with models with large parameters and significant sampling steps. Furthermore, the diffusion architecture poses challenges for seamless integration with LLMs and limit their potential in unified multimodal systems.

In this paper, we aim to bridge the diffusion process and autoregressive modeling for visual generation, leveraging strengths from both paradigms.
Crucially, we maintain a strict adherence to the standard next-token prediction paradigm, without any modifications to the underlying autoregressive mechanism.
To achieve this, we present the coarse-to-fine {\em sequential diffusion tokenizer} to reinterpret the diffusion process on raw image pixels as a sequence of discrete tokens.
In this formulation, early tokens represent conditions in early diffusion steps from pure noise, whilst later tokens capture progressive steps over less noised inputs, leading to a naturally linearized decomposition of visual sequence~\cite{heatdiffusion}.
We design the diffusion model in the proposed diffusion tokenizer to be light and fast, e.g., with around 185M parameters and 8 diffusion steps, and achieve 1.52 rFID on ImageNet~\cite{imagenet} with a total budget of 256 discrete tokens.
With this design, we can perform the diffusion process on image pixels via predicting next token in token sequence with the autoregressive mechanism untouched.
Therefore, we name this framework as {\textbf{D}ffusion {v}ia \textbf{A}uto\textbf{r}egressive models} (\textbf{D-AR}).

D-AR offers several intriguing properties inherited from both diffusion and autoregressive worlds, including \textbf{1)} it natively supports fast inference with KV cache, \textbf{2)} it provides consistent previews when partial tokens are generated, and \textbf{3)} it can be easily extended to zero-shot layout-controlled synthesis by conditioning several prefix tokens.
Besides these interesting properties, D-AR also excels on the standard ImageNet class-conditioned generation benchmark.
With a 775M-parameter, LLaMA backbone~\cite{touvron2023llamaopenefficientfoundation} operating purely in the next-token-prediction paradigm, D-AR-XL achieves 2.09 gFID with a total of 256 tokens.
We hope our work can inspire future research on integrated multi-modal LLM architectures with native visual generation capabilities.

\begin{figure}[t]
  \vspace{-1em}
  \centering
  \includegraphics[width=1.0\textwidth]{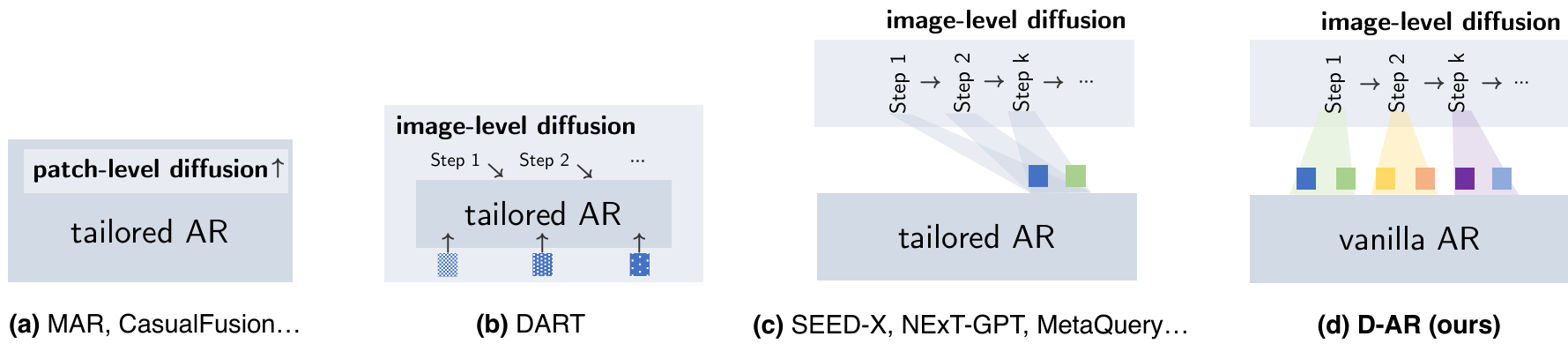}
  \caption{\textbf{Different paradigms incorporating diffusion and autoregressive models} for vision generation. (a) uses patch-level diffusion during every single autoregressive step to tackle continuous outputs~\cite{mar,causalfusion,diffusionforcing}; (b) DART~\cite{dart} denoises a full image per every autoregressive step by AR transformer, together inputted with history denoised images; (c) use a single set of continuous outputs to different diffusion steps and require diffusion gradients to train AR models~\cite{dreamllm,seedx,nextgpt,metaquery}; (d) ours uses vanilla AR models, which can train with discrete inputs/outputs by simply cross entropy, and sequentially decode output tokens with our diffusion tokenizer.}
\end{figure}
\section{Related Work}
\label{related_work}
\subsection{Diffusion and Autoregressive Models}
Diffusion models and autoregressive models are currently two main streams of modern generative modeling.
Diffusion models~\cite{ddpm,ddim,flowmatching,straightflow}, exemplified by several commercial text-to-image models~\cite{flux,dalle}, excel in generating high-quality visual content by iteratively denoising a sample from an initial noise.
Though powerful in generating visually pleasing images, the diffusion process typically operates in a dense manner and requires significant sampling steps, which can be computationally expensive.
Recent success in language modeling using autoregressive paradigm, especially large language models~\cite{grattafiori2024llama,openai2024gpt4technicalreport,gemini,qwen}, has inspired researchers to explore the potential of this paradigm in visual generation tasks due to its scalability and mature training and inference infrastructures.
However, this adaptation raises several challenges, since images are not inherently discrete and linear structures like text.
To this end, researchers use vector quantized autoencoders to quantize images into discrete latent codes~\cite{vqvae,vqgan} and use raster-scan order to model the image sequence~\cite{llamagen, chameleon}.
Researchers have also found that image sequence ordering can be defined in various ways~\cite{var,randar,xar,rar}, and the next-token prediction paradigm should be adapted to suit vision modeling accordingly.

Though sorts of visual autoregressive models have been proposed, the dominant role of diffusion models in visual generation tasks remains almost unchanged due to their outperforming capabilities at visual continuous signals.
In this paper, we seek to bridge diffusion models and autoregressive models for visual generation and leverage the advantage of both sides, following previous efforts in this research line~\cite{mar,dart,diffusionforcing,causalfusion,nextgpt,metaquery,seedx,transfusion,far}. But different from these work, we strictly adhere to the standard next-token-prediction autoregressive paradigm with discrete inputs and outputs, and design diffusion in the tokenizer decoder in a sequential manner to tackle with visual continuous data.

\subsection{Visual Tokenization with Diffusion Models}
How to encode images into sequences of discrete tokens and then effectively reconstruct pixels from them is a key design for visual generation in autoregressive models.
Due to the vector quantization and downsampling operations, visual tokenization methods inevitably suffer from the loss of information and lead to suboptimal reconstruction quality, which researchers have put intensive efforts into improving~\cite{vqvae,vqgan,improvedvqgan,rqtransformer}.
Concurrently, a research direction recently emerges on leveraging diffusion models to decode visual tokens back into image pixels~\cite{openai_consistencydecoder_2023,epsilonvae,hart,flomo,hart}.
Specifically, these methods typically see discrete tokens as conditions in the diffusion process.
By doing so, they offload visual ambiguity and fine details to the diffusion model and significantly improves the visual fidelity~\cite{epsilonvae,dito,flomo}.
Further work on this line argues that discrete tokens should focus on structural semantics of images and extract such semantics with flexible sequence length~\cite{semanticist, flextok} by large latent diffusion models together with VAE~\cite{vae,ldm}.

To our best knowledge, our method is the first to propose the tokenizer to interpret the full diffusion process into the autoregressive sequential generation using the diffusion tokenizer. Our method is individually developed from a recent conceptually related work, DDT-LLaMa~\cite{ddtllama}, which also uses a diffusion decoder to sequentialize tokens but in a reversed order.
As a consequence, notably, DDT-LLaMa cannot represent diffusion steps as sequential AR generation process and therefore cannot decode with partial tokens generated by autoregressive models, marking a key distinction from our method and underlying motivation.

\section{Methods}
\label{methods}
There is a recent debate in the vision community on whether autoregressive modeling surpasses diffusion models in visual generation tasks~\cite{llamagen,var}, given the great success of large language models.
A critical challenge in autoregressive modeling is, for a long time, how to tokenize a 2D image into a sequence of discrete tokens since images are not inherently 1D linear structures like text.
Though several works defined the ordering of image pixels~\cite{randar,var,rar,xar}, they either introduce spatial inductive bias or require tailored autoregressive designs for vision, posing challenges on a unified autoregressive framework.

We propose a systematic solution to address this by introducing {Diffusion via Autoregressive models} ({D-AR}), which recasts the image diffusion process as a fully autoregressive model in the standard next-token-prediction manner.
The high-level idea is to \textit{perform the diffusion process on pixels via autoregressive modeling}.
To start, we design a \textit{sequential diffusion tokenizer} that tokenizes images into sequences of 1D discrete tokens, which can be sequentially decoded as diffusion steps from the first to the end token.
We apply standard next-token prediction on these tokens using a Llama-like autoregressive model~\cite{touvron2023llamaopenefficientfoundation}, without modifying any AR architecture (either causal masks or training/inference designs) to generate images.

\subsection{Sequential Diffusion Tokenizer}
\begin{figure}[t!]
    \centering
     \includegraphics[width=1.0\textwidth]{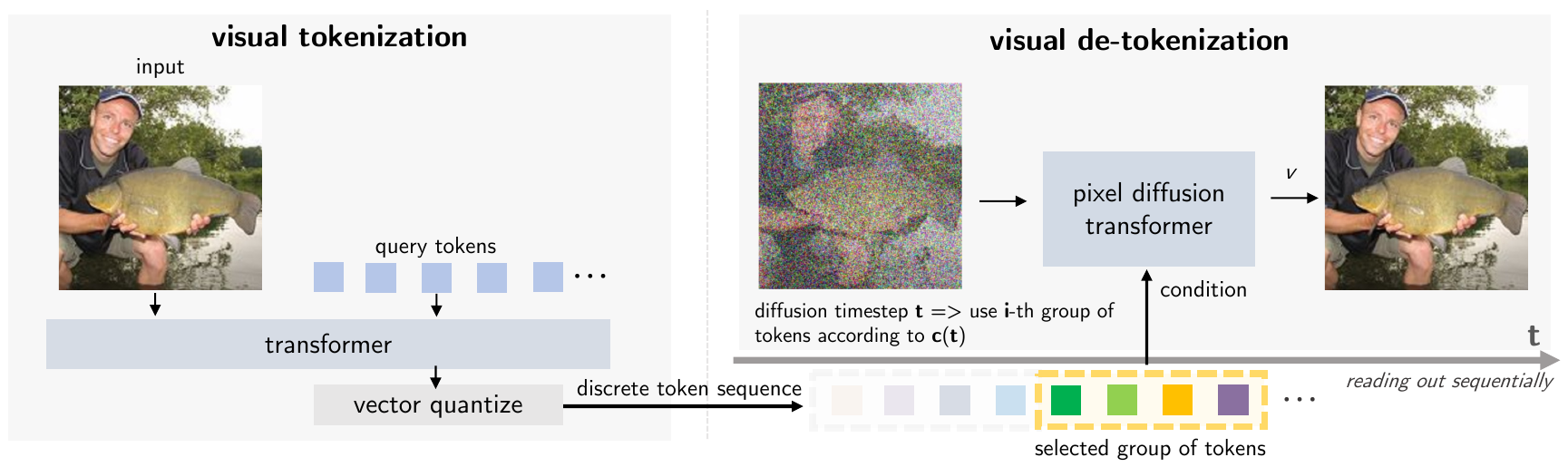}
    \caption{\textbf{Sequential diffusion tokenizer structure.} When training the tokenizer, the pixel diffusion transformer in the tokenizer decoder calculates the velocity loss with the selected group of tokens, $\mathbf{c}(t)$, as conditioning tokens.}
    \label{fig:tokenizer}
\end{figure}
\label{sec:diffusion_tokenizer}
The {sequential diffusion tokenizer} is designed to tokenize images into 1D linearized sequences of discrete tokens in the ordering of progressive diffusion steps. The overall tokenizer structure is shown in Figure~\ref{fig:tokenizer}, akin to conventional visual tokenizers, which encodes images into latents, quantize them into discrete ones, and then decode them back into diffusion over pixels in an auto-encoding manner.
\paragraph{1D encoding.}
Similar to 1D tokenization approach~\cite{titok}, the sequential diffusion tokenizer first encodes the image into a 1D sequence of discrete tokens using a transformer:
\begin{align}
    \mathbf{z} = [\mathbf{z}_1, \mathbf{z}_2, \ldots, \mathbf{z}_N] &= \text{quant}(\mathcal{E}(\mathbf{I}, [\mathbf{q}_1, \mathbf{q}_2, \ldots, \mathbf{q}_N])),
\end{align}
where $\mathbf{I}$ is the input image, typically patchified as a set of patch tokens, $\mathcal{E}$ is the transformer encoder~\cite{transformers}, $\text{quant}(\cdot)$ is the vector quantizer~\cite{vqvae}, and $[\mathbf{q}_i]$ are learnable query tokens, where $N$ is the total number of queries.  In this step, we do not impose a specific ordering on the resulting 1D token sequence, which we will further focus on below.

\paragraph{Sequential diffusion decoding.}
We propose the sequential diffusion decoder to decode 1D quantized token sequence into consecutive diffusion steps on image pixels. The diffusion decoder is a diffusion transformer~\cite{dit}, which takes tokens in different positions in the sequence as conditions in different diffusion steps. Here, flow matching loss with velocity prediction, a simplified variant of diffusion families~\cite{straightflow,flowmatching,sit}, is used to train the diffusion decoder. The loss is defined as:
\begin{align}\label{eq:flow_matching}
    \ell_{\text{fm}} &= \mathbb{E}_{t, \mathbf{x}_0, \mathbf{x}_1} \left[ \left\| \mathbf{v}_t - \mathcal{D}_{\text{FM}}(\mathbf{x}_t, t, \mathbf{c}(t)) \right\|_2^2 \right],
\end{align}
where the flow interpolant is defined as:
\begin{align}
    \mathbf{x}_t = t\mathbf{x}_1 + (1-t)\mathbf{x}_0, \hspace{1em} &\mathbf{v}_t=d \mathbf{x}_t / d t=\mathbf{x}_1-\mathbf{x}_0,\\
    \mathbf{x}_0\sim \mathcal{N}(0, 1), \hspace{1em} &\mathbf{x}_1=\mathbf{I}, \hspace{1em} t\in [0, 1].
\end{align}
In this notation, $\mathbf{x}_0$ at timestep $t=0$ represents pure noise and $\mathbf{x}_1=\mathbf{I}$ at $t=1$ represents the real data sample. During inference, samples can be generated by solving ordinary differential equation (ODE) from $t=0$ to $t=1$ when the condition schedule $\mathbf{c}(t)$ is given.

The condition schedule $\mathbf{c}(t)$ is a set of quantized tokens $\mathbf{z}_i$ used as conditions in the diffusion decoder at timestep $t$.
To enable the sequential decoding property, we design the condition schedule $\mathbf{c}(t)$ to start from the first token $\mathbf{z}_1$ and reach the last token $\mathbf{z}_N$ as the flow matching timestep $t$ progresses from $0$ to $1$.
In preliminary experiments, we find that multiple $\mathbf{z}_i$ for a specified timestep is crucial for good performance. We thus first group consecutive tokens $\mathbf{z}_i$ into $K$ groups, $\{\mathbf{g}_1, \mathbf{g}_2, \ldots, \mathbf{g}_K\}$, each group $\mathbf{g}_i$ with $N/K$ tokens. The condition schedule is then defined as:
\begin{align}\label{eq:condition_schedule}
    \mathbf{c}(t) &=\mathbf{g}_{\lceil t' \cdot K \rceil}, \hspace{2em} t'
    ={t}/{(t+(1/\beta) * (1-t))},
\end{align}
where $t'$ is the shifted timestep and $\beta$ is a control parameter. When $\beta=1$, time ranges are evenly split regarding the condition group $\mathbf{g}_i$. The higher $\beta$ values lead to denser tokens as conditions over early diffusion steps, which we find empirically beneficial for reconstruction quality.

\paragraph{Discussion.}
One can view the 1D sequence of tokens as the ``proxy'' of the underlying diffusion procedure on pixels controlled by conditioning tokens $\mathbf{c}(t)$.
With sequential diffusion decoding, we can decode increments of AR tokens into consecutive diffusion sampling steps on pixels in the streaming way when reading out tokens sequentially.
The ordering of tokens is naturally linearized by the diffusion process, where early tokens represent conditions needed in early diffusion steps ($t\rightarrow0$) over noisy inputs, often low-frequency global semantics or spatial layout.
Later tokens describe the information needed in later steps ($t \rightarrow 1$) over less noisy inputs, typically localized details or fine-grained structures~\cite{heatdiffusion}.
We argue that this coarse-to-fine, linearized token ordering is well-suited for autoregressive modeling.
Also, by the diffusion decoder, the tokenizer decoder can delegate ambiguous details to diffusion and thus focus on semantics~\cite{hudson2024soda}.

\subsection{Autoregressive Modeling}
\label{sec:autoregressive_modeling}
Once we have the linearized sequence of discrete tokens by our proposed tokenizer, we can apply standard autoregressive next token prediction to model the image generation process:
\begin{align}
    p_\theta(\mathbf{z}) &= \prod_{i=1}^N p_\theta(\mathbf{z}_i | \mathbf{z}_1, \ldots, \mathbf{z}_{i-1}),
\end{align}
where $\theta$ is the AR model parameters and one can use simple cross entropy loss to optimize parameters. In this paper, we resort to the decoder-only transformer architecture~\cite{touvron2023llamaopenefficientfoundation,llamagen} for autoregressive modeling.

\paragraph{Vanilla vision autoregressive modeling.}
General autoregressive modeling assumes a linearized order of data elements, which is hard to define in images.
Thanks to tokens linearized by the sequential diffusion tokenizer, we do not need to modify any component in the underlying autoregressive modeling.
That is, D-AR models do not change discrete inputs and outputs, customize attention masks or kernels, or vary training loss function/inference logistics.
In contrast, most visual autoregressive models require tailoring the autoregressive mechanism specialized for vision and carefully re-design the underlying components of decoder-only transformers.

\subsection{Diffusion via Autoregressive Models}
\label{sec:diva}
The presented framework, diffusion via autoregressive models, simply consists of the sequential diffusion visual tokenizer and the Llama decoder-only transformer on discrete token sequences. The sequential diffusion tokenizer directly operates on raw pixels and do not require extra VAEs~\cite{vae,ldm}.

\paragraph{Markovian diffusion procedure via vanilla autoregressive models.}
As the name implies, sequential generation in the D-AR framework directly corresponds to diffusion procedure on image pixels via the bridge of token conditioning. When we are generating a sequence of tokens, we can perform the diffusion sampling on pixels simultaneously whenever we have condition tokens needed at diffusion timestep $t$ ready, i.e., $\mathbf{c}(t)$.
Since the diffusion is only controlled by autoregressive models via condition tokens, we do not break the Markovian convention of diffusion models, different from~\cite{dart}.
Therefore, D-AR can leverage advantages of both diffusion and autoregressive sides:
\begin{enumerate}
    \item \textbf{KV cache-friendly inference:} as the D-AR framework uses autoregressive decoder-only transformers on token sequences, it natively supports KV cache-friendly fast inference;
    \item \textbf{Streaming pixel decoding and consistent previews at no extra costs.} We can perform diffusion steps on pixels instantly whenever we have needed tokens ready in a streaming manner. Also, since the diffusion decoder is directly operating on pixels, we can use the diffusion property to jump-estimate the target and generate consistent previews effortlessly;
    \item \textbf{Zero-shot controlled synthesis.} As the token sequence is linearized by diffusion, we can simply condition several prefix tokens to control the visual generation without finetuning.
\end{enumerate}

\section{Implementations}
\label{sec:implementation}
\paragraph{Sequential diffusion tokenizer architecture.}
For the encoder in diffusion tokenizer, we mainly follow the design of 1D tokenizer~\cite{titok} to use the transformer encoder layers jointly processing image patches and learnable query tokens. We apply a causal mask to query tokens to enforce the basic causality on queries but allow both query tokens and image tokens to attend to arbitrary image tokens. As default, we set the number of queries $N=256$, input patch size $p=16$, the dimension of transformer $d=768$, and the transformer layer $L=8$.
Following the practice of~\cite{llamagen}, we use the vanilla vector quantization with $\ell_2$-normalized codebook entries, configured with codebook size $n_e=16384$ and dimension $d_e=8$. We expect better performance with more advanced quantization approaches~\cite{fsq,lfq}, but we leave for future work.

We design the diffusion decoder as the diffusion transformer architecture~\cite{dit,sit} but on raw pixel patches, which integrates zero-initialized adaptive layer normalization (AdaLN)\cite{adaln}.
To condition the diffusion decoder with condition tokens $\mathbf{c}(t)$, we use the cross attention layer on patch tokens to attend to condition tokens and take attention output as the input of the AdaLN, together added by the time $t$ embedding. The diffusion transformer decoder is configured moderately with $L_d=12$ layers, $d_d=768$ hidden dimension, and patch size $p_d=8$, resulting in a total parameter of 185M.

We add causal decoder transformer layers on encoded tokens $\mathbf{z}$, after the vector quantization and before diffusion decoding, to produce $\mathbf{z}'$ for more nonlinearity. We configure it as the same as the transformer encoder. Note that these decoder transformer layers with causal masks do not break the causality of the token sequence. The total parameter of the sequential diffusion tokenizer is 300M.
\paragraph{Training sequential diffusion tokenizer.}
Training diffusion models on raw pixels with few inference steps is a challenging task~\cite{simplediff,simplerdiff}, even with the strong image encoded conditions~\cite{epsilonvae}. To enable few-step inference and speed up the convergence, we use the perceptual matching loss based on LPIPS~\cite{lpips,epsilonvae} and representation alignment (REPA) loss~\cite{repa} together with flow matching (\ref{eq:flow_matching}) and vector quantization loss to train the sequential diffusion tokenizer:
\begin{align}
    \ell_{\text{tokenizer}}=\ell_{\text{fm}}+\ell_{\text{VQ}}+\lambda_1 \ell_{\text{LPIPS}}+\lambda_2 \ell_{\text{repa}},
\end{align}
 where we assign $\lambda_1=0.5$ and $\lambda_2=0.5$. We do not use adversarial matching loss~\cite{epsilonvae} in our training since we observe the instability and over-saturation issue.

In a training forward pass, we first encode an image into a quantized token sequence and use transformer decoder layers to compute $\mathbf{z}'$.
Then we randomly sample a flow matching timestep $t\in[0, 1]$, determine which group $\mathbf{g}_i$ of $\mathbf{z}'$ should be used as conditions for diffusion decoder according to the condition schedule $\mathbf{c}(t)$, and compute the final loss $\ell_\text{tokenizer}$.

\paragraph{Sampling with sequential diffusion tokenizer.}
Given the token sequence, either encoded from images or generated from autoregressive modeling, we can perform the flow matching sampling by reading out tokens in the sequential order based on the condition schedule $\mathbf{c}(t)$.
For simplicity and efficiency, we design the default sampling schedule to use each condition group exactly once, that is, to bind the number of sampling steps to the number of condition groups $K$ and use a timeshifted schedule in the reversed form of (\ref{eq:condition_schedule}), following~\cite{sd3}:
\begin{align}
    t_i = \frac{i/K}{(i/K)+\beta * (1-i/K)}, \hspace{1em} i=0, 1, \ldots, K-1.
\end{align}
This sampling schedule results in denser early sampling steps when $\beta>1$ and we default set $\beta=2$ and $K=8$ for sampling efficiency, resulting in each conditioning group with $N/K=32$ tokens. Again, for efficiency, we do not use classifier-free guidance (CFG)~\cite{cfg} in diffusion sampling steps.

\paragraph{AR models.}
Our AR model architecture is exactly the same as Llama decoder-only transformers, which are equipped with RMSNorm~\cite{rmsnorm} and SwiGLU~\cite{swisglu}.
Note that since tokens by sequential diffusion tokenizer are inherently one-dimensional, we apply the original 1D RoPE~\cite{rope}, rather than 2D RoPE, in attention layers as positional embedding.
The class conditions, e.g., image labels, are injected as a single prefix token following~\cite{llamagen}.
We do not use AdaLN in our AR models.
Classifier-free guidance on logits is used during AR inference.
We mainly design two variants of D-AR models, D-AR-L and D-AR-XL, with 343M and 775M parameters respectively, also following~\cite{llamagen}.

To generate an image, D-AR models first produce a sequence of tokens conditioned on the given label in the standard token-by-token manner with KV cache enabled.
In pace with sequential generation, we can decode tokens generated into diffusion sampling steps on pixels either concurrently or offline.

\section{Experiments}
\label{experiments}
\paragraph{Experimental Setup.}
We conduct D-AR experiments on the ImageNet $256\times 256$ class-conditional generation benchmark~\cite{imagenet}. The sequential diffusion tokenizer is trained on the ImageNet training set with a batch size of 1024, Adam optimizer~\cite{adam} of learning rate $2*10^{-4}$ and a total of 210K iterations till convergence, together with an exponential moving average with a $0.999$ decay rate. The training procedure took around 5 days on 16 A100 GPUs to finish.
We follow the training recipe~\cite{randar} to train D-AR autoregressive models with a batch size of 1024 for 300 epochs. We use AdamW optimizer~\cite{adamw} with learning rate $4*10^{-4}$, $(\beta_1, \beta_2)=(0.9, 0.95)$ and weight decay of $0.05$.
The learning rate is decayed to $1*10^{-5}$ linearly within the last 50 epochs, following \cite{randar}. It took about 2 and 3 days on 16 A100 GPUs to finish training D-AR-L and D-AR-XL with 343M and 775M parameters respectively.
The performance of D-AR is evaluated in terms of FID~\cite{fid}, Inception Score~\cite{inceptionscore}, precision and recall scores, following the standard ADM evaluation pipeline~\cite{adm}. For the reconstruction performance of the sequential diffusion tokenizer, we mainly investigate the reconstruction FID (rFID) on the ImageNet validation 50K set.
\subsection{Main Results}
\label{sec:main_results}
\paragraph{Tokenizer results.}
\TableTokenizerComparison
We investigate the key component of our D-AR framework, i.e., the sequential diffusion tokenizer.
In Table~\ref{tab:tokenizer_comparison}, we compare our sequential diffusion tokenizer with the conventional LlamaGen tokenizer, which has the same budget of 256 tokens and the same vector quantization configuration, as strong baselines.
Despite having more parameters (300M versus 72M), which is mainly due to the pixel diffusion decoder, our sequential diffusion tokenizer achieves better reconstruction fidelity and is more endurable to smaller codebook size.

\TableTokenizerSampling
\TableSystemComparison
We also study different sampling configurations of the proposed sequential diffusion tokenizer in Table~\ref{tab:tokenizer_sampling}, where we vary the sampling steps and flow matching ODE solver. We resort to the two-step Adams–Bashforth solver for flow matching with 8 steps as it provides clearer samples without increasing numbers of function evaluations (NFEs) on the diffusion decoder.

\paragraph{System-level comparison.}To compare with state-of-the-art methods, we experiment with D-AR models on the ImageNet $256\times 256$ class-conditional generation benchmark. Following~\cite{randar}, the linear CFG schedule is used in D-AR (1.1$\to$8.0 for D-AR-L and 1.1$\to$10.0 for D-AR-XL).
In Table \ref{tab:system_comparison}, D-AR-L and D-AR-XL achieve the leading level of performance in their parameter count regions.
Among vanilla AR models in the strict token-by-token fashion, D-AR-XL achieves 2.09 FID with 775M parameters, outperforming LlamaGen-XXL and competing with IBQ-XXL 2.1B.

\begin{figure}[t]
    \centering
    \includegraphics[width=0.95\textwidth]{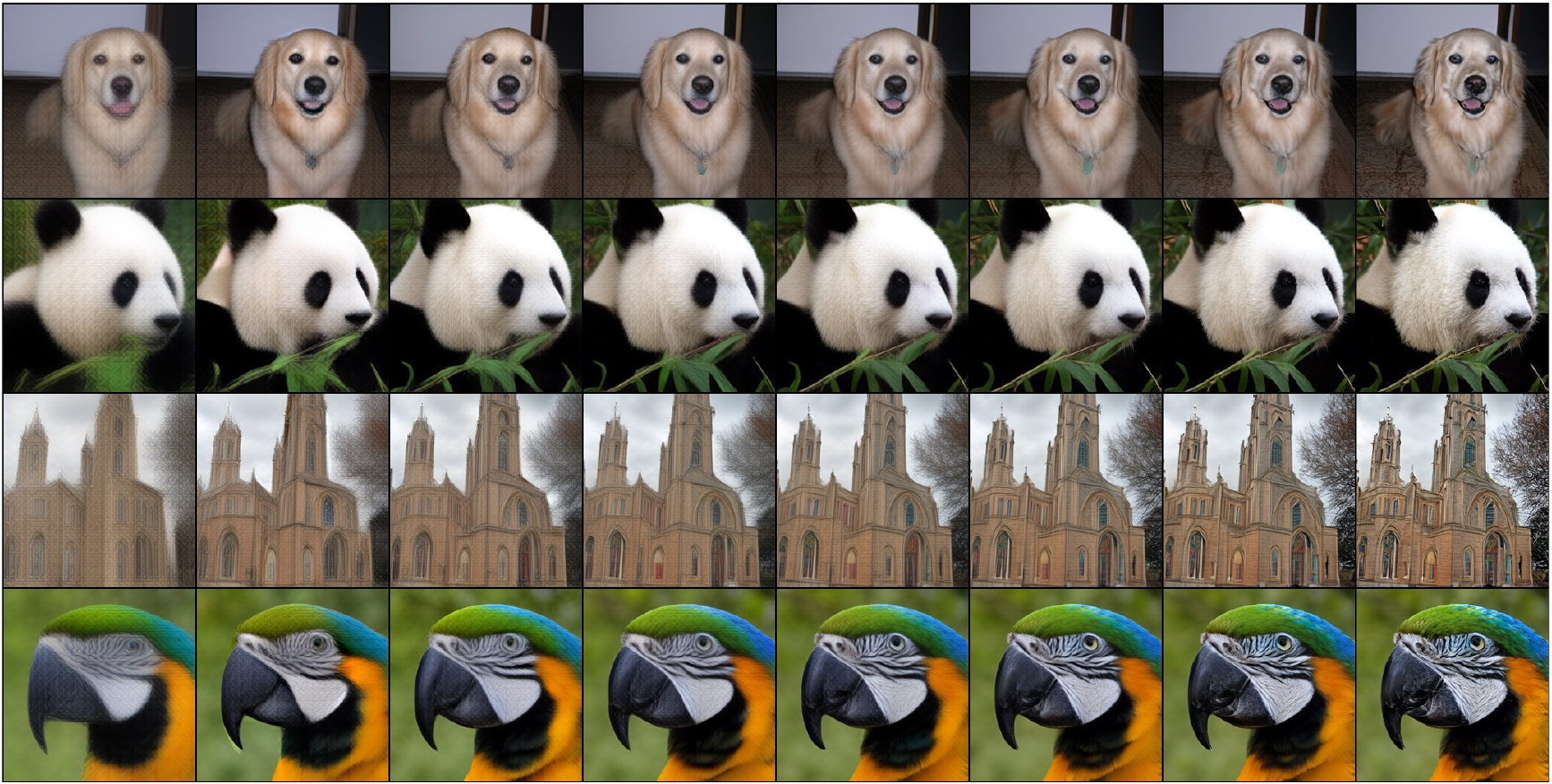}
    \caption{\textbf{Consistent previews as generation trajectories} for every increment of 32 tokens (a group). Note that these previews can be generated in a streaming manner with AR tokens partially generated.}
    \label{fig:preview}
\end{figure}

Recent attempts to incorporate diffusion into autoregressive models, such as CausalFusion, DART-FM, and MAR, have also shown highly competitive results.
However, they require significant modifications in the autoregressive framework to tackle continuous-valued inputs and outputs of images.
In contrast, D-AR maintains the vanilla autoregressive mechanism with favored performance.
\paragraph{Consistent previews and generation trajectories.} As discussed in Section~\ref{sec:diffusion_tokenizer}, the sequential diffusion tokenizer can generate consistent previews of generated images when partial tokens are generated, inherited from the diffusion property to jump-estimate the target $\hat{\mathbf{x}}_1=(1-t)\mathbf{v}_t+\mathbf{x}_t$ for every sampling timestep $t$.
As our diffusion model is on raw pixels, this operation takes almost no extra cost. 
We visualize these previews in Figure~\ref{fig:preview}, which are consistent with final samples.
These previews can also be interpreted as generation trajectories of our autoregressive model, which inherently follow a coarse-to-fine progression, in line with~\cite{heatdiffusion}.

\begin{figure}[t]
    \centering
    \includegraphics[width=0.99\textwidth]{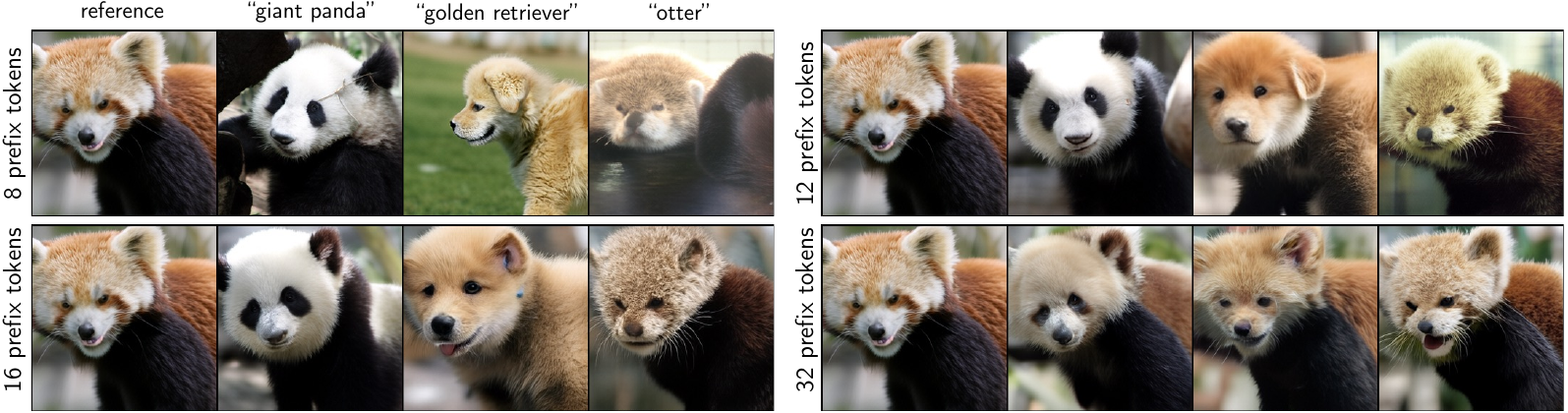}
    \caption{\textbf{Zero-shot layout-controlled synthesis} with different prefix tokens and varying labels.}
    \label{fig:layout}
\end{figure}

\paragraph{Zero-shot layout-controlled synthesis.}
We also investigate the zero-shot layout-controlled synthesis with D-AR, where several prefix tokens are given and fixed, in Fig~\ref{fig:layout}.
Thanks to the linearized structure by the diffusion decoder, we can generate plausible images with reference layouts conditioned on reference prefix tokens and given labels, without specific finetuning.
As more prefix tokens are provided, layout control becomes stronger, while label-relevant information increasingly concentrates on fine-grained details such as fur textures. 
We include more ablation studies and qualitative results in the appendix.

\section{Conclusion}
\label{conclusion}
\begin{wrapfigure}{r}{0.4\textwidth} 
\vspace{-5em}
\centering
\includegraphics[width=0.38\textwidth]{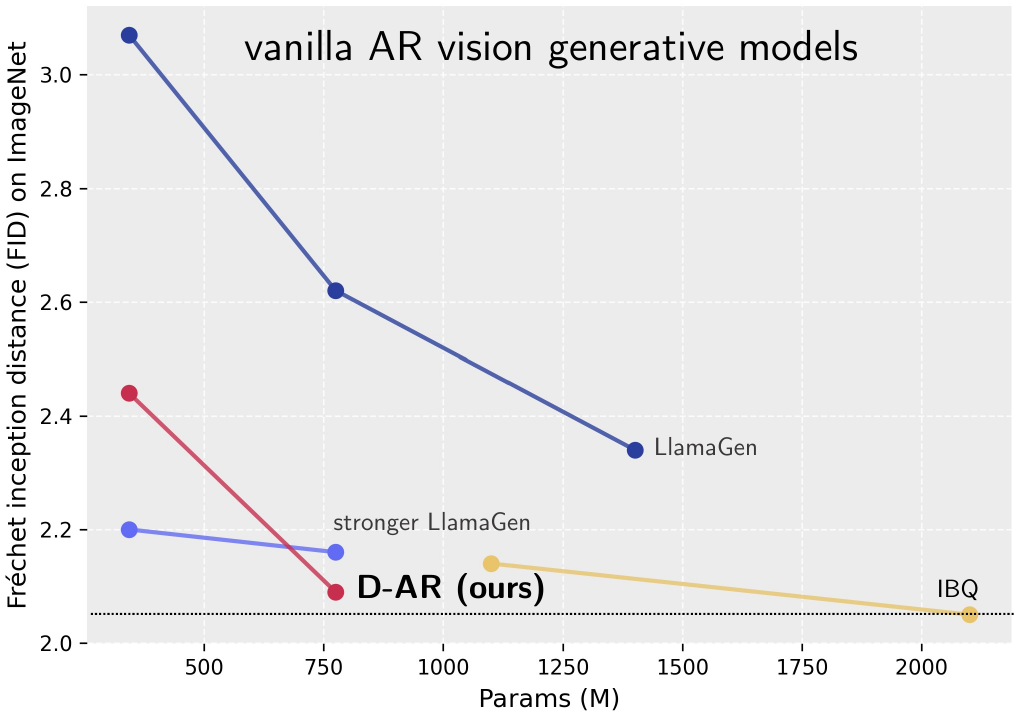} 
\caption{Vanilla AR comparison for ImageNet generation.}
\vspace{-2em}
\end{wrapfigure}
In this paper, we present Diffusion via Autoregressive models (D-AR), a framework to bridge the pixel diffusion and autoregressive modeling for visual generation.
With the linearized sequence of discrete tokens by the presented sequential diffusion tokenizer, we can perform vanilla autoregressive process in the standard next token prediction fashion.
Thus, the AR sequence generation process in the D-AR framework directly mirrors consecutive diffusion denoising steps on pixels.
Experiments on the standard ImageNet $256\times 256$ benchmark shows that D-AR can generate high-quality images, reaching competitive 2.09 FID with 775M parameters as a vanilla autoregressive model, together with several properties from both autoregressive and diffusion worlds.

\paragraph{Limitation.}
Due to the limited hardware resources, we only experiment with D-AR models of a relatively moderate amount of parameters (<1.0B) on ImageNet.
Though our method is generally designed for unified autoregressive architectures for native visual generation compatible with LLMs, we do not explore the native text-to-image generation in this paper, where yet we believe the D-AR framework holds significant potential values.

\paragraph{Acknowledgment.}
This research is supported by the National Research Foundation, Singapore under its AI Singapore Programme (AISG Award No: AISG3-RP-2022-030).
We would like to thank Zhan Tong, Tong He, and Shuai Wang for helpful discussions and comments.

\section*{I. Detailed Architecture of Sequential Diffusion Tokenizers}
\paragraph{Vector quantization.}
We follow LLamaGen~\cite{llamagen} to set up the vanilla vector quantization~\cite{vqvae} as well as its loss: $\ell_{\text{VQ}} = ||\text{sg}[f] - z||_2^2 + \beta ||f - \text{sg}[z]||_2^2$, where $\text{sg}[\cdot]$ is the stop gradient operator and $\beta=0.25$. We do not impose the entropy loss on codebook learning.
\paragraph{Transformer architecture.} In our sequential diffusion tokenizer, we adopt the transformer architecture with vanilla LayerNorm~\cite{layernorm} and SiLU activation function~\cite{silu}. We also apply QK normalization~\cite{qknorm} in attention computation for training stability. For tokens with explicit spatial locations, e.g., those patchified from images in the tokenizer encoder or in diffusion transformer, we apply the 2D RoPE~\cite{rope} in attention to encode spatial relations. For those who do not have 2D inherent locations, i.e., 1D query tokens in the transformer encoder and decoder, we simply disable rotation in RoPE by using the identity matrix.

\section*{II. Detailed Evaluation of D-AR Models }
\begin{table}[h!]
  \centering
    \setlength{\tabcolsep}{4.4pt}
    \renewcommand{\arraystretch}{1.1}
    \caption{D-AR with different CFG schedules. The `none' indicates disabling CFG.}\label{tab:detailedcfg}
    \begin{tabular}{c|c|cccc}
        model & CFG schedule & FID$\downarrow$ & IS$\uparrow$ & Prec$\uparrow$ & Recall$\uparrow$\\
        \shline
        D-AR-L & none  & 7.43 & 117.60 & 0.71 & 0.63 \\
               & 1.5   & 3.50 & 245.22 & 0.83 & 0.54 \\
               & 1.75  & 4.70 & 291.76 & 0.86 & 0.50 \\
               & 1.1$\to$8.0 & 2.44 & 262.97 & 0.78 & 0.61 \\
        \hline
        D-AR-XL & none & 5.11 & 145.78 & 0.73 & 0.64 \\
        & 1.5 & 3.39 & 276.37 & 0.84 & 0.55 \\
        & 1.1$\to$10.0 & 2.09 & 298.42 & 0.78 & 0.62
    \end{tabular}
\end{table}
\paragraph{CFG schedules.}
In the main paper, we present the performance of D-AR-L and D-AR-XL with linear CFG schedule, following RandAR~\cite{randar}. Note that previous work also explore customized CFG schedule for better performance, as a common practice on ImageNet~\cite{mar,randar,rar}.  We report D-AR models results with different CFG strategies in Table~\ref{tab:detailedcfg}. The models here are exactly the models in the main paper in Table 3. We do not use top-p, top-k, and temperature in our sampling in the main paper and appendix.

\begin{table}[h]
  \centering
    \setlength{\tabcolsep}{8.4pt}
    \renewcommand{\arraystretch}{1.1}
    \caption{D-AR-L jump-estimation results with partial AR tokens.}\label{tab:jumpestimation}
    \begin{tabular}{c|cccc}
        \#AR tokens
        & 64  & 128 & 192 & 256 \\
        \#diffusion steps
        & 2  & 4  & 6 & 8 \\
        \shline
        FID$\downarrow$
        & 7.38 & 3.94 & 2.93 & 2.44        \\
        IS$\uparrow$ 
        & 165.25 & 227.74 & 257.08 & 262.97 \\
        Prec$\uparrow$ 
        & 0.74 & 0.78 & 0.80 & 0.78 \\
        Recall$\uparrow$
        & 0.48 & 0.54 & 0.57 & 0.61
        
    \end{tabular}
\end{table}

\paragraph{Partial AR tokens results.}
In the main paper, we have visualized the diffusion target sample estimation with partial AR tokens generated. We here report quantitative results by D-AR-L in Table~\ref{tab:jumpestimation}.

\section*{III. Tokenizer Ablations}
Due to the limited computation resource, we design a lightweight version of our proposed sequential diffusion tokenizer with 113M parameters (we change the dimension in the transformer to 512 and the depth of diffusion transformer to 8) and train for 50K iterations with 256 batch size. This ablation training typically takes about 8 hours to complete on 4 A100s.

\begin{table}[h]
  \centering
    \setlength{\tabcolsep}{8.4pt}
    \renewcommand{\arraystretch}{1.1}
    \caption{Effects of $\beta$ on tokenizer training.}\label{tab:beta}
    \begin{tabular}{c|ccc}
        $\beta$
        & 1  & 2 & 4 \\
        \shline
        rFID$\downarrow$
        &  39.75 & 28.65 & 27.10 \\
        codebook utilization$\uparrow$ 
        & 97.8 & 99.4 & 99.8
        
    \end{tabular}
\end{table}
\paragraph{Ablations on $\beta$ in conditioning schedules.} The control parameter, $\beta$, in the conditioning schedule, also acts as timeshift parameter in the diffusion procedure in our sequential diffusion tokenizer. Since we are operating on pixels, we set $\beta$ to 2 as default. Here we investigate different $\beta$ on tokenizer training in Table~\ref{tab:beta}. We can see that there is a large gap between $\beta=1$ and $\beta=2$ in reconstruction FID as well as in coodebook utilization, while $\beta=2$ and $\beta=4$ matches closer. These empirical results show that early diffusion need denser steps as well as AR tokens as conditioning on diffusion in the pixel space.

\begin{table}[h]
  \centering
    \setlength{\tabcolsep}{8.4pt}
    \renewcommand{\arraystretch}{1.1}
    \caption{Effects of $K$ on tokenizer training. For fair comparison, for $K<8$ variants, we use 8 diffusion sampling steps to decode images.}\label{tab:kk}
    \begin{tabular}{c|cccc}
        $K$
        & 1  & 4 & 8 & 16 \\
        \shline
        rFID$\downarrow$
        &  23.18 & 23.04 & 28.65 & 49.07 \\
        codebook utilization$\uparrow$ 
        & 99.0 & 99.5 & 99.4 & 91.5 \\
        gFID$\downarrow$ & 52.91 & 53.70 & 54.26 & 63.69
        
    \end{tabular}
\end{table}
\paragraph{The numbers of conditioning group $K$.} The number of conditioning group $K$ decides how many tokens are feed into pixel diffusion model per diffusion step, $N/K$. We investigate the effects of $K$ in the Table~\ref{tab:kk}. The sequential diffusion tokenizer with single group $K=1$ with multiple sampling steps degrades into conventional tokenizers with diffusion decoder~\cite{flomo,dito}, which denoises an image with full token sequence on every timestep. This $K=1$ setup  does not yield a linearized ordering of visual tokens and lacks the sequential nature central to our approach.

For reconstruction FID here, it is reasonable for small group number variants to perform better, since the number of conditioning tokens per denoising step become more as $K$ decreases. The linearized order also becomes weak as $K$ approaches $1$. In the other side, $K=16$ enforces the diffusion-induced linearized order most strongly, but came out with the worst reconstruction FID.
We also train a D-AR-B with 111M parameters for 50 epochs with tokens by these tokenizers.
Though the reconstruction FID with $K=8$ lags behind $K=4$ variant, the generation FID achieved by the D-AR-B models remains comparable. This suggests that more strongly linearized token sequences may facilitate faster convergence for autoregressive models. Therefore, we choose $K=8$ as our default choice to balance trade-off between reconstruction quality and linearized structure for generation.

\section*{IV. More Visualizations}
\begin{figure}[h]
  \centering
  \includegraphics[width=1.0\textwidth]{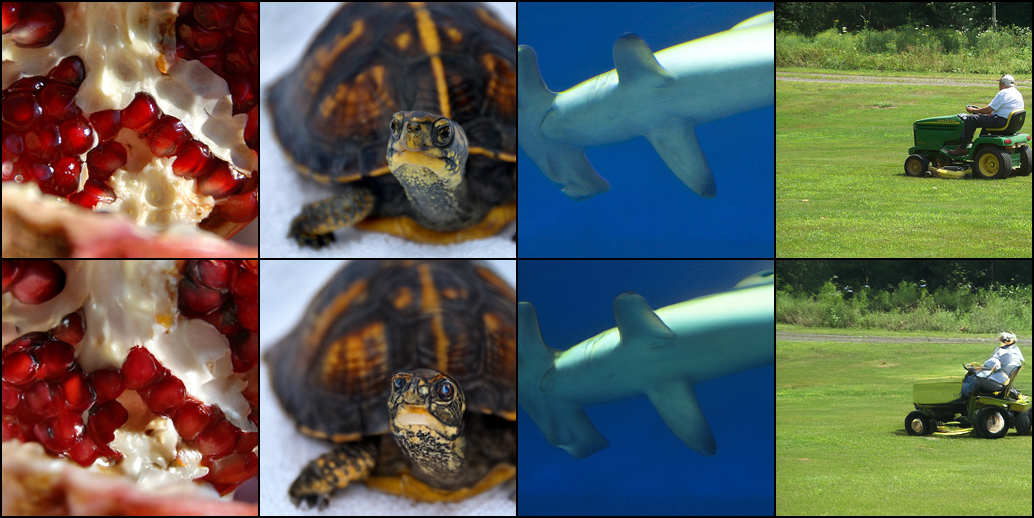}
  \includegraphics[width=1.0\textwidth]{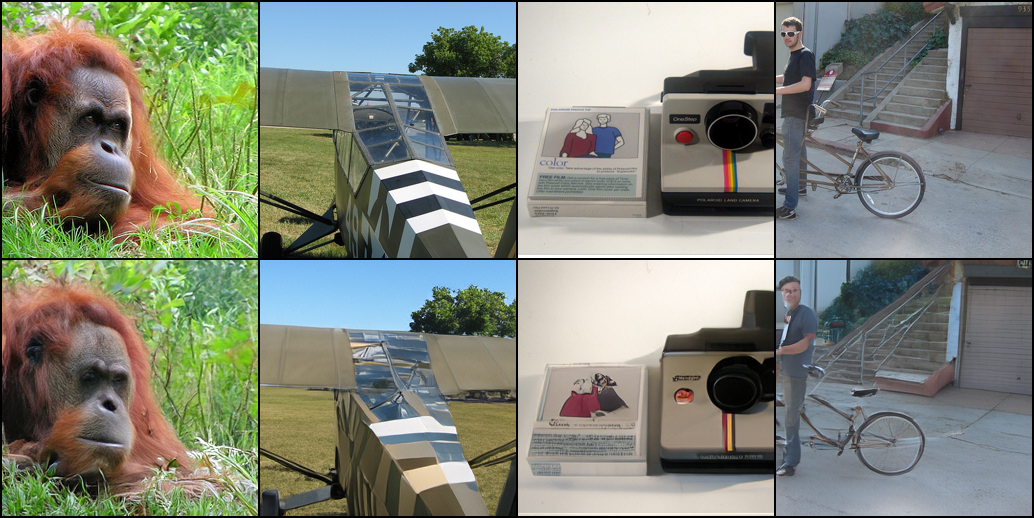}
  \includegraphics[width=1.0\textwidth]{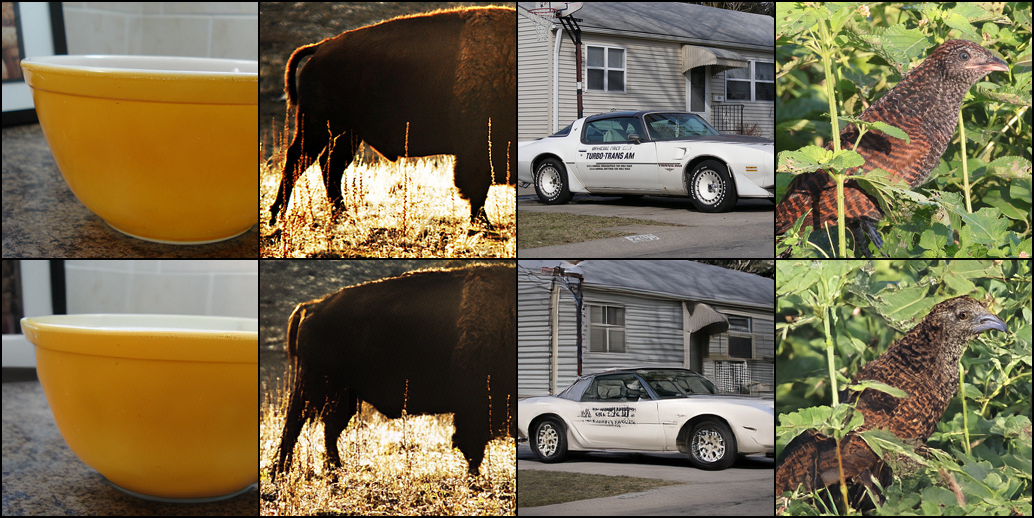}
  \caption{\textbf{Reconstruction results} with samples from the ImageNet validation set. Each pair of rows shows: first row — input; second row — reconstruction.}\label{fig:recon}
\end{figure}

\paragraph{Tokenizer reconstruction results.} We also present reconstruction samples from our sequential diffusion tokenizer (rFID = 1.52) in Fig~\ref{fig:recon}. As observed, fine details are not strictly reconstructed, which is mainly attributed to the inherent stochastic and denoising nature of the diffusion process.
Since our primary objective is to model image generation rather than achieve exact pixel-level reconstruction, this trade-off is acceptable and consistent with our diffusion tokenizer design.

\begin{figure}[b]
  \centering
  \includegraphics[width=1.0\textwidth]{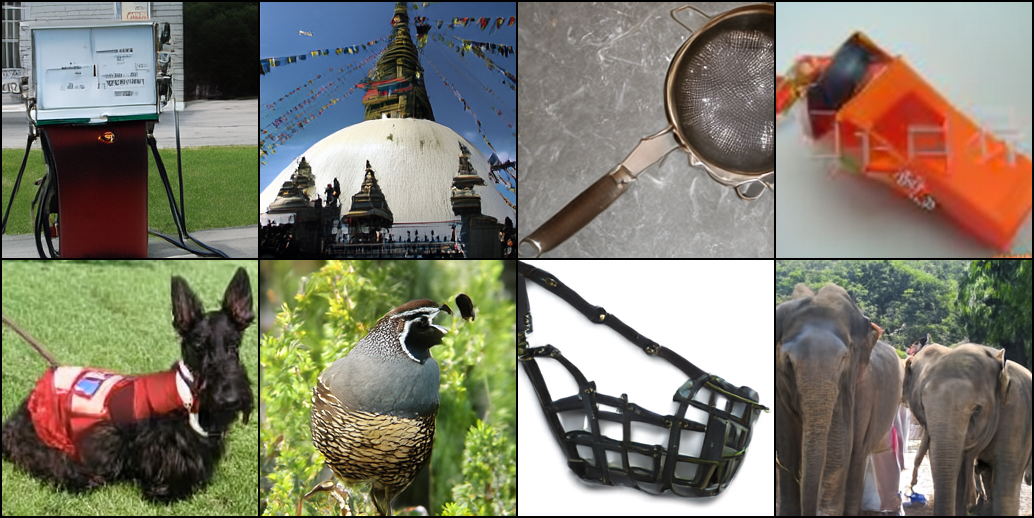}
  \caption{\textbf{Uncurated generated samples} by D-AR-XL with random labels and CFG=4.0.}\label{fig:sample}
\end{figure}

\begin{figure}[h]
  \centering
  \includegraphics[width=1.0\textwidth]{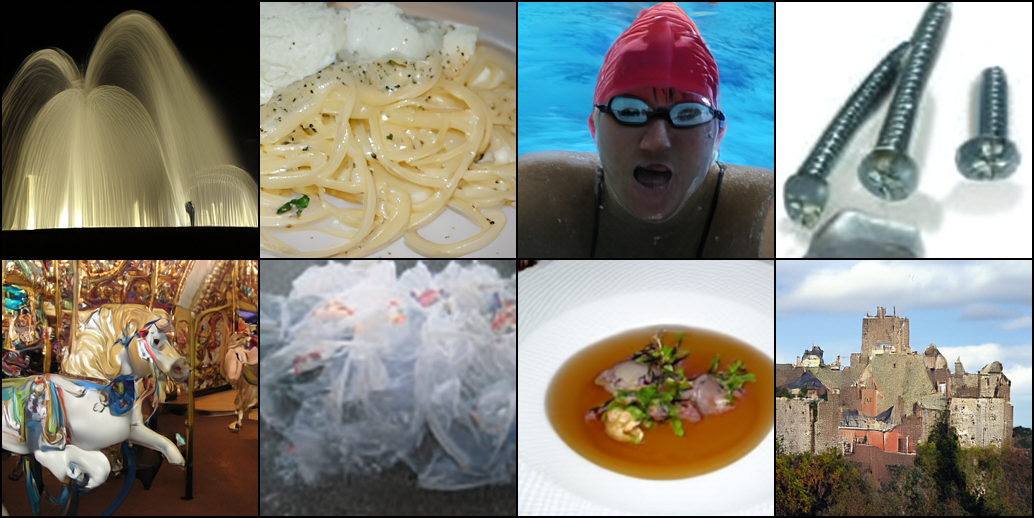}
  \includegraphics[width=1.0\textwidth]{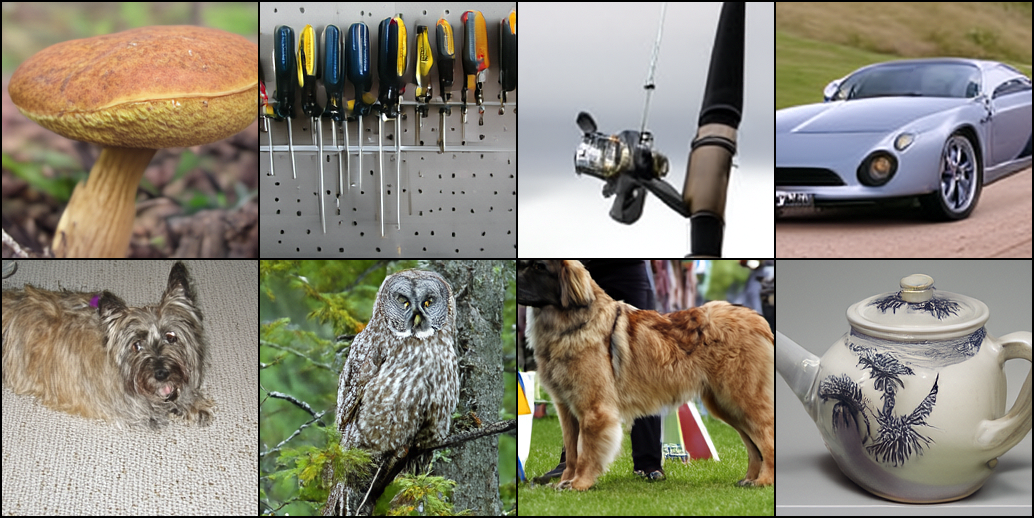}
  \includegraphics[width=1.0\textwidth]{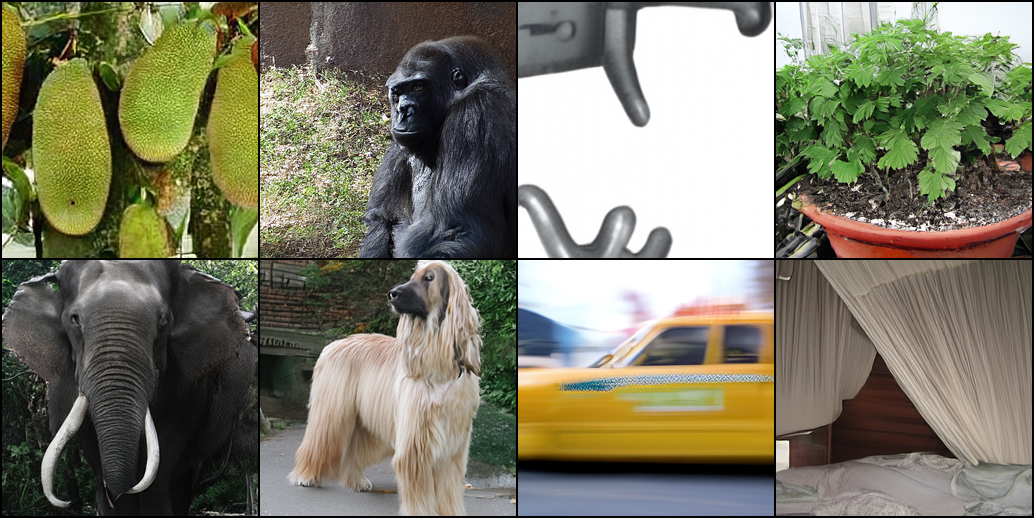}
  \caption{\textbf{Uncurated generated samples} by D-AR-XL with random labels and CFG=4.0 (cont'd).}\label{fig:sample2}
\end{figure}

\begin{figure}[t]
  \centering
  \includegraphics[width=1.0\textwidth]{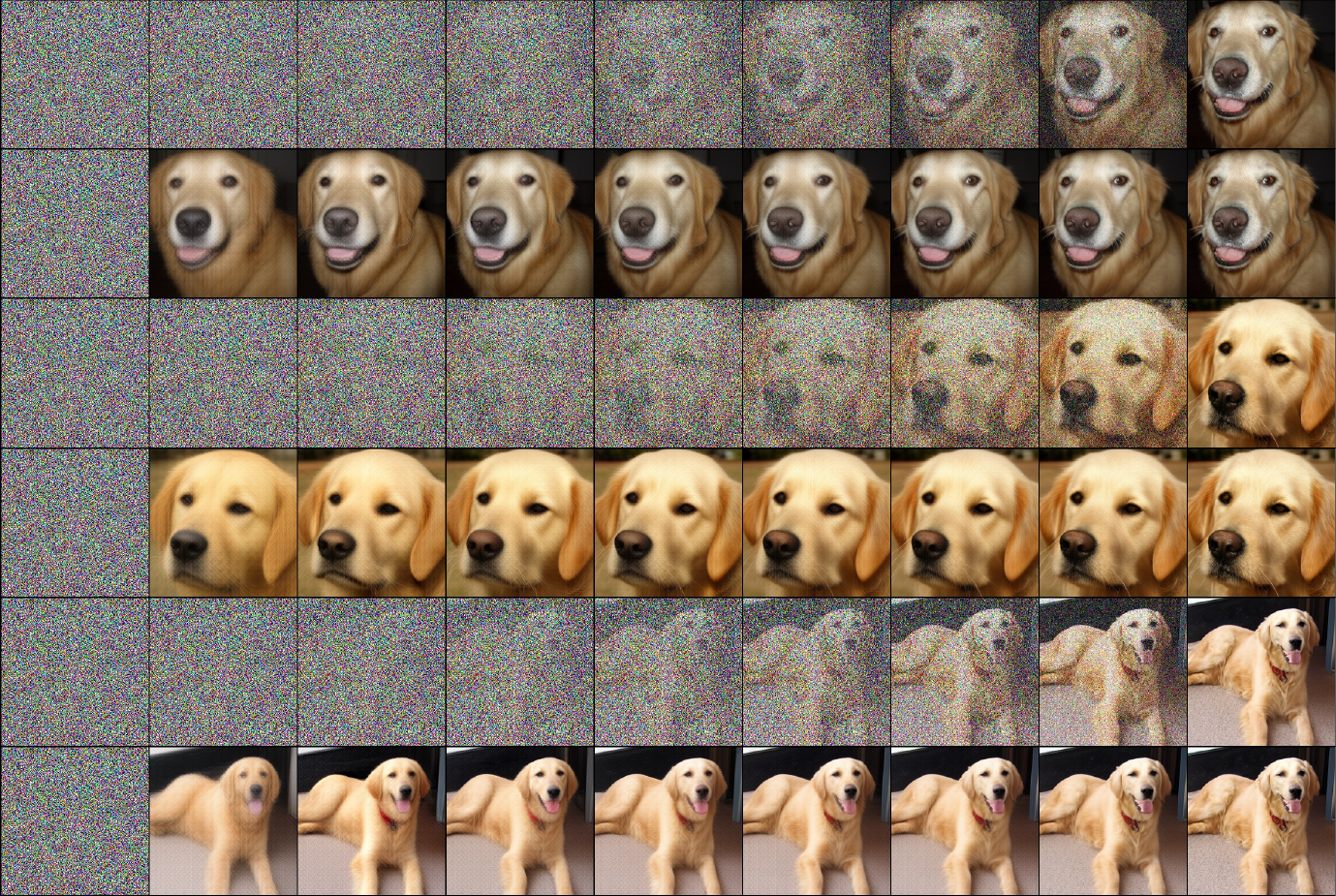}
  \includegraphics[width=1.0\textwidth]{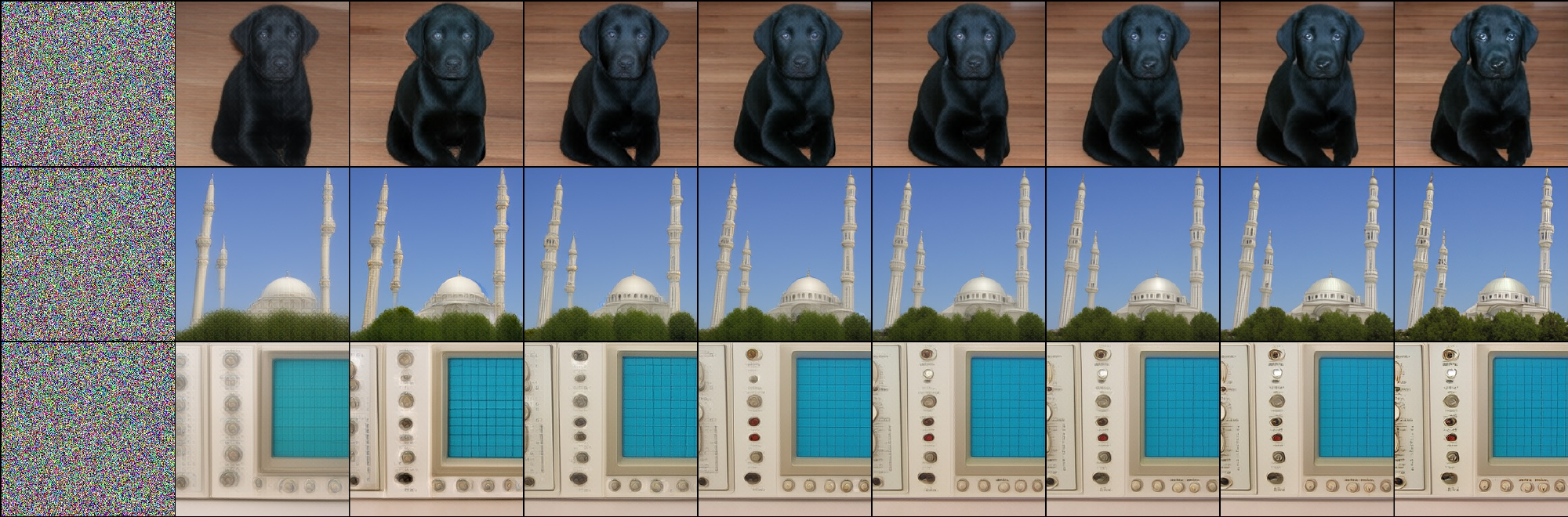}
  \caption{\textbf{Generation trajectory and previews} at each diffusion sampling step by D-AR-L.}\label{fig:traj}
\end{figure}

\paragraph{Generation trajectories.} We show more generation trajectories as well as previews in Fig~\ref{fig:traj}. Our D-AR models follow coarse-to-fine generation with consistent previews with final targets.

\begin{figure}[t]
  \centering
  \includegraphics[width=1.0\textwidth]{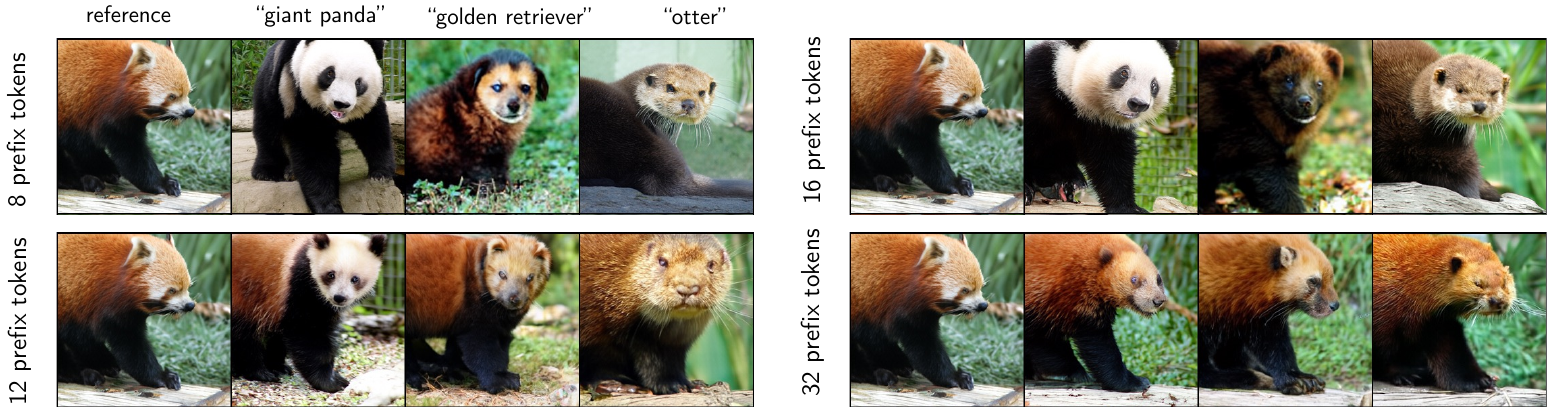}
  \caption{Zero-shot layout-controlled synthesis.}\label{fig:zeroshot2}
\end{figure}

\paragraph{Zero-shot layout-controlled synthesis.} As discussed in the main paper, we can simply condition on prefix tokens to generate layout-following images in a zero-shot manner. We here show more zero-shot layout-controlled generated samples by fixing different numbers of prefix tokens and varying labels in Fig~\ref{fig:zeroshot2}. 

\clearpage
{
\small
\bibliography{mainbib}
\bibliographystyle{unsrt}
}

\end{document}